\documentclass[10pt,twocolumn,letterpaper]{article}

\usepackage[pagenumbers]{cvpr} % To force page numbers, e.g. for an arXiv version

\usepackage{float}
\usepackage{graphicx}
\usepackage{xspace}
\usepackage{longtable}
\usepackage{algpseudocode}
\usepackage{listings}
\usepackage[ruled]{algorithm2e}
\usepackage{multirow}
\definecolor{cvprblue}{rgb}{0.21,0.49,0.74}
\usepackage[pagebackref,breaklinks,colorlinks,allcolors=cvprblue]{hyperref}

\def\paperID{5594} % *** Enter the Paper ID here
\def\confName{CVPR}
\def\confYear{2025}

\title{Enhancing Image Matting in Real-World Scenes with Mask-Guided Iterative Refinement}

\author{Rui Liu\\
{\tt\small ruiliu.new@gmail.com}
}

\newif\ifversionlatest
\versionlatesttrue 
\begin{document}
\maketitle
% inputversion based on version selection
%
    \ifversionlatest
        \begin{abstract}
Real-world image matting is essential for applications in content creation and augmented reality. However, it remains challenging due to the complex nature of scenes and the scarcity of high-quality datasets. To address these limitations, we introduce Mask2Alpha, an iterative refinement framework designed to enhance semantic comprehension, instance awareness, and  fine-detail recovery in image matting. Our framework leverages self-supervised Vision Transformer features as semantic priors, strengthening contextual understanding in complex scenarios. To further improve instance differentiation, we implement a mask-guided feature selection module, enabling precise targeting of objects in multi-instance settings. Additionally, a sparse convolution-based optimization scheme allows Mask2Alpha to recover high-resolution details through progressive refinement,from low-resolution semantic passes to high-resolution sparse reconstructions. Benchmarking across various real-world datasets, Mask2Alpha consistently achieves state-of-the-art results, showcasing its effectiveness in accurate and efficient image matting. 
\end{abstract} % Input latest version
    \else
        \input{sec/0_abstract_v2} % Input old version with suffix
    \fi

    \ifversionlatest
        %-------------------------v2 version------------------------------
\section{Introduction}
\label{sec:intro}
\begin{figure}[t]
    \centering
    \includegraphics[width=0.9\linewidth]{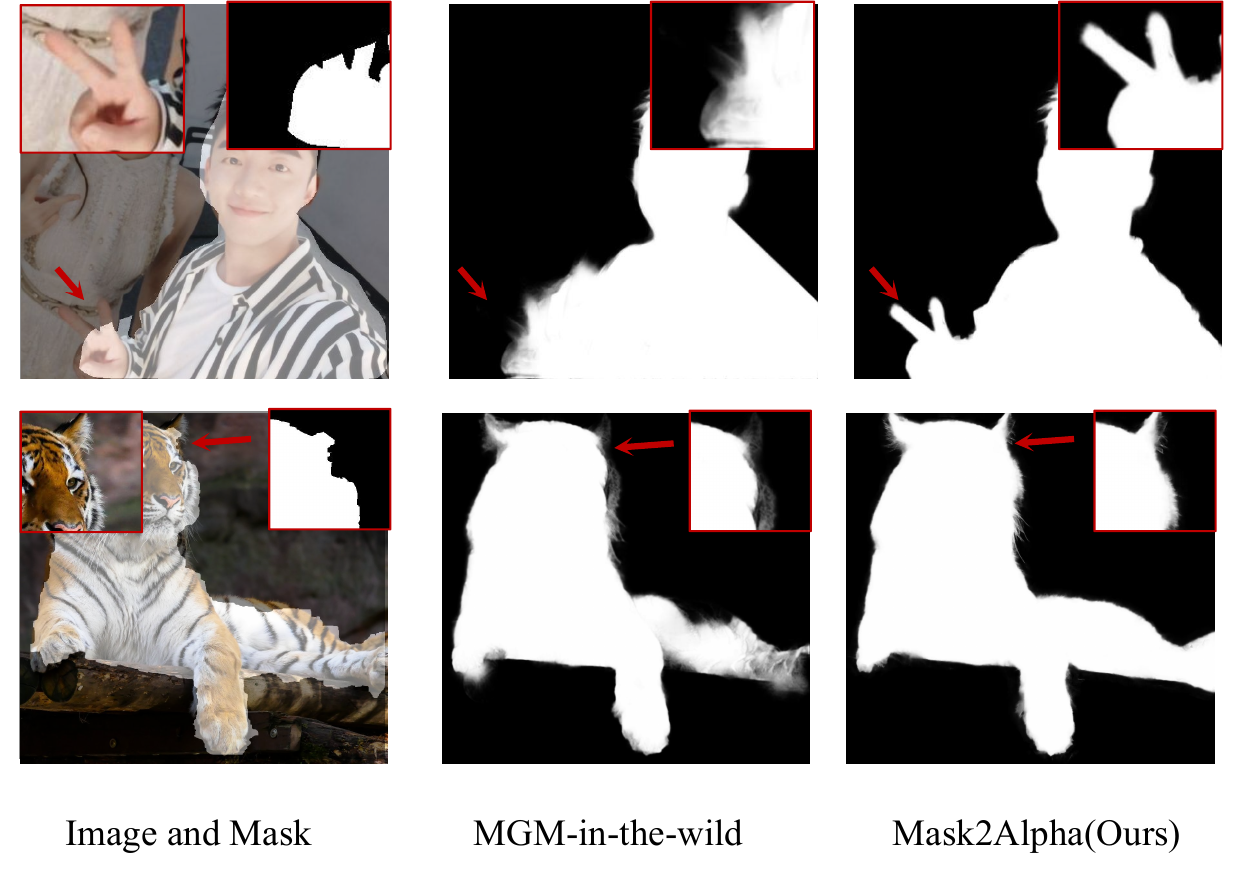}
    \caption {MGM-in-the-wild\cite{park2023mgmwild} often fail in real-world applications, particularly when handling fine-grained object details and reducing edge errors. We propose \textbf{Mask2Alpha} to address the difficulties of real-world scenarios.}
    \label{fig:teaser}
\end{figure}

\begin{figure*}[t]
    \centering \includegraphics[width=0.8\linewidth]{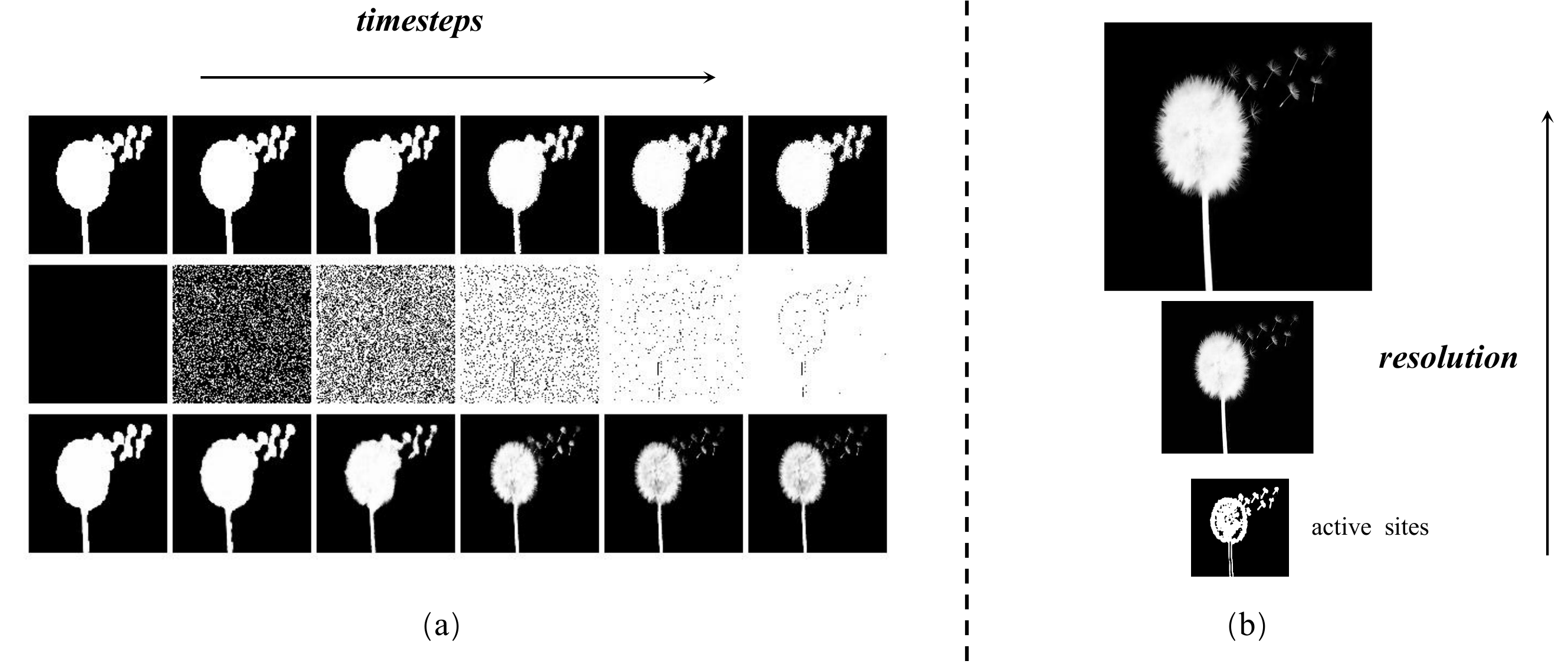}
    \caption{\textbf{Iterative Optimization Process.} The Mask2Alpha framework operates in two stages: (a) Semantic Iterative Optimization - begins by refining high-confidence regions through a state transition matrix, where the first row represents the input mask, the second row displays the state transition, and the third row shows the resulting semantic output; (b) Detail Iterative Optimization - progressively enhances uncertain fine details following semantic refinement, aiming to recover the optimal solution across varying resolutions.}
    \label{fig:iterate}
\end{figure*}

Image matting is a fundamental problem in computer vision, aiming to separate foreground objects from the background through accurate alpha matte estimation. Traditional methods typically rely on user-defined trimaps~\cite{wang2007optimized, Sun2004PoissonM, aksoy2017designing}, which help reduce uncertainty by clearly delineating foreground, background, and unknown regions. Other approaches have incorporated scribbles~\cite{zheng2008fuzzymatte, wang2005iterative} or background knowledge~\cite{backgroundmatting, backgroundmattingv2} to assist in matting, while recent works have increasingly focused on using automatically generated segmentation masks~\cite{park2023mgmwild, yu2021mgm, huynh2024maggie}. Following this trend, we also leverage segmentation masks as key auxiliary information in our approach. 

Despite these advances, real-world image matting remains challenging due to the complexity of scenes, semantic ambiguity, and the need for high-resolution processing. As shown in \cref{fig:teaser}, real-world images often suffer from a lack of precise semantic information, particularly in regions where object boundaries are unclear, leading to errors in foreground-background separation. This is further compounded by the challenge of recovering fine details, which are essential for accurate matting but difficult to extract in complex scenes. Inspired by iterative image generation techniques~\cite{maskgit}, we propose a novel iterative framework, Mask2Alpha, which progressively refines alpha matte predictions. As shown in the process illustrated in \cref{fig:iterate}, our method begins by focusing on high-confidence regions and iteratively corrects low-confidence areas, thereby improving the overall accuracy of the matte. This iterative approach allows our model to better handle the inherent complexity of real-world scenes and recover fine details more effectively.

Another major challenge is the scarcity of large, diverse datasets that accurately capture the complexity of real-world scenes, limiting the generalization capabilities of existing matting models. Most available datasets focus on specific object categories or simple scenes, leaving a gap in the ability of models to generalize to more complex and diverse environments. To address this, we leverage visual foundation models(VFMs), which have shown great promise in capturing rich, semantic representations of images without the need for extensive labeled data~\cite{caron2021emerging}. By incorporating a mask-guided feature enhancement module, our approach refines the semantic understanding of the scene, enabling the model to focus on relevant features of the target instance while filtering out irrelevant background or noise. This explicit guidance improves the accuracy of alpha matte computation, even in the most challenging real-world settings.

Lastly, handling high-resolution images is crucial for preserving fine details, but it also brings substantial computational challenges, especially when iterative refinement is required. High-resolution image matting requires intense focus on fine details from high-resolution data, while other regions often introduce significant computational redundancy. Methods like SparseMat~\cite{sun2023sparsemat} leverage sparse convolution to enhance high-resolution processing. However, these methods rely on dilating low-resolution results to select sparse regions, which can lead to inefficient thresholds that either increase computational load or compromise detail retention. To address this, we propose a self-guided sparse region selection strategy that dynamically identifies key regions for refinement without morphological operations. This approach optimizes computational efficiency while preserving high-quality detail.

In summary, Mask2Alpha combines mask-guided input, VFMs feature enhancement, and a coarse-to-fine optimization strategy to tackle the complex challenges of matting in real-world environments. By addressing issues such as boundary confusion, dataset limitations, and high-resolution processing, we present a novel framework that advances the state-of-the-art in image matting.

Our contributions can be summarized as follow:

\begin{itemize} \item[$\bullet$] We propose a mask-guided image encoder, which leverages high-level contextual information from visual foundation models, resulting in enhanced matting quality in semantically complex scenes.

\item[$\bullet$] We propose an iterative refinement framework that progressively enhances alpha matte predictions by initially focusing on high-confidence regions, effectively improving detail in edge-confused areas where mask boundaries are ambiguous.

\item[$\bullet$] We introduce a self-guided sparse detail recovery module that dynamically targets key areas for refinement, ensuring efficient high-resolution processing and precise detail recovery. 
\end{itemize}

 % Input latest version
    \else
        \input{sec/1_intro_v2} % Input old version with suffix
    \fi

    \ifversionlatest
        \section{Related Work}\label{sec:related_work}

\vspace{-3pt}
\subsection{Image matting}

Image matting seeks to extract foreground objects from images, typically using a trimap to indicate foreground, background, and unknown regions. Traditional methods can be categorized as sampling-based, which estimate alpha mattes using color samples from known regions, or propagation-based, which transfer alpha values based on pixel affinity. However, both approaches often struggle with complex scenarios due to their reliance on low-level color features. 

The advent of deep learning has led to significant advancements in matting techniques. Methods like DIM utilize convolutional encoder-decoder architectures and sophisticated loss designs, improving performance through attention mechanisms. Some approaches aim to eliminate trimap dependency, such as using additional background images or user interactions, but often fail to generalize well to unseen objects in real-world settings. 

Recent developments have introduced mask-guided matting frameworks that only require coarse masks for guidance. MGMatting~\cite{yu2021mgm} has been a pioneer in this area, while MGM-in-the-wild~\cite{park2023mgmwild} explores how to train a generalized mask-guided matting model, often without delving deeply into the model's inner workings. InstMatt~\cite{sun2022instmatt} addresses the challenge of accurately matting overlapping human instances with intricate boundaries, enabling effective separation in complex scenes.MaGGIe~\cite{huynh2024maggie} and SparseMat~\cite{sun2023sparsemat} focuses on efficient, end-to-end instance matting to enhance practical applicability across diverse categories. In contrast, our method investigates the complexities of mask-based challenges in real-world environments. We aim to construct an adaptable framework for mask-guided matting that builds upon the foundational aspects of the model, thereby improving performance across a wide range of object types and challenging backgrounds.

\subsection{Segmentation And Matting Refinement}

In natural image matting, refining high-frequency details around object boundaries is crucial for achieving visually appealing results. Traditional segmentation refinement techniques, like PointRend, enhance masks using convolutional networks and MLPs, relying on coarse predictions from Mask R-CNN and point-wise confidence scores to identify uncertain regions, but they suffer from limited adaptability due to their dependence on manually tuned hyperparameters. In contrast, EFormer employs a semantic and contour detector with a cascade of cross-attention and self-attention mechanisms, facilitating interaction between local details and global semantics for improved detail localization. Model-agnostic strategies, such as SegFix, aim to refine masks across various models but depend heavily on accurate object detection, particularly in complex datasets. SegRefiner, based on a discrete diffusion process, addresses boundary errors but is hindered by slow processing due to its iterative nature, making it less suitable for real-time applications. Non-autoregressive transformers, exemplified by MaskGIT~\cite{maskgit}, offer a more efficient solution by quantizing images into discrete tokens and utilizing parallel decoding, significantly enhancing processing speed. By leveraging these models, we can effectively capture detailed features and maintain coherent semantic relationships in complex scenes.

\subsection{Self-Supervised Learning}
Self-supervised learning (SSL) has become a powerful framework for extracting effective feature representations without human annotations, leveraging pretext tasks to derive supervision directly from data. Deep models, particularly self-supervised vision transformers (DINO and DINOv2~\cite{DINOv2}), have proven effective for dense correspondence under varying photometric and geometric changes. DINOv2, an enhanced version of DINO, shows strong generalization across tasks like classification and segmentation, although its application to correspondence tasks is still underexplored. In image matting, approaches like ViTMatte\cite{Yao2024vitmatte} adapt ViTs for improved feature extraction, while MatAny and MAM utilize the SAM model as a prior. SMat harnesses the rich semantics from ViT features and creatively employs the class token as a saliency cue for guiding salient object localization. By leveraging self-supervised representations, our approach aims to enhance the precision and efficiency of image matting, highlighting the potential of SSL in this domain.

 % Input latest version
    \else
        \input{sec/2_related_work_v2} % Input old version with suffix
    \fi

    \ifversionlatest
        \section{Mask2Alpha}
\begin{figure*}[t]
    \centering \includegraphics[width=0.8\linewidth]{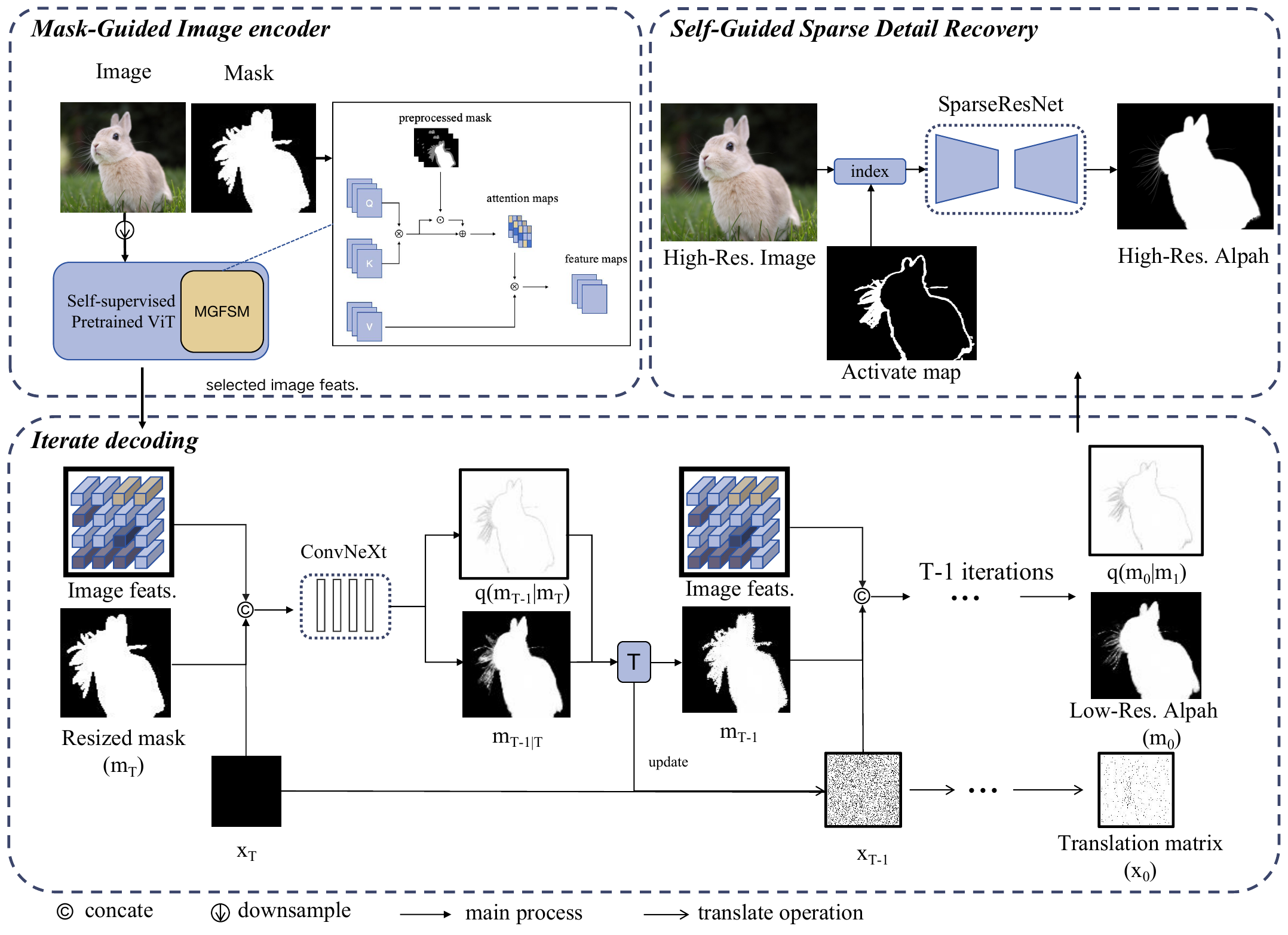}
    \caption{\textbf{The Pipeline of our Mask2Alpha.}The process begins with the Input Image and Initial Mask, which are processed by the Mask-Guided Image Encoder to extract multi-scale features guided by semantic regions. These features are then passed to the Iterative Decoding stage, where alpha mattes are progressively refined over multiple iterations. Finally, the  Self-Guided Sparse Detail Recovery stage uses adaptive fusion with confidence-weighted feature maps to output the final refined alpha matte with enhanced high-resolution detail and precision.}
    \label{fig:frame}
\end{figure*}

Our framework, illustrated in \cref{fig:frame}, processes an image \( I \in [0, 255]^{3 \times H \times W} \) and a binary guidance mask \( M \in \{0, 1\}^{H \times W} \), producing an alpha matte \( A \in [0, 1]^{H\times W} \). The pipeline consists of three main stages: (1) Mask-Guided Image Encoder (Section~\ref{sec:encoder}): The input image and mask are resized to \( H // 2 \times W // 2 \) and passed through the encoder to extract multi-scale features. (2) Iterate Decoding (Section~\ref{sec:iter}): This stage takes the multi-scale image features and a resized \( H // 4 \times W // 4 \) mask as input, and iteratively refines the output to generate a low-resolution alpha matte along with a confidence map. (3) Self-Guided Sparse Detail Recovery (Section~\ref{sec:decoder}): The decoder uses the low-resolution alpha and the guidance map generated by the confidence map to produce the final high-resolution alpha matte at \( H \times W \) resolution. Through these stages, our method progressively refines alpha matte predictions, improving both detail and accuracy at higher resolutions.

Finally, Section~\ref{sec:procedure} outlines our training and inference process, focusing on the iterative refinement mechanism and strategies. 

% Section~\ref{sec:explore} explores key challenges encountered in mask-guided matting within natural scenes, presenting experimental insights that underscore the need for our approach. Section~\ref{sec:architecture} details the architecture, elaborating on the components that drive Mask2Alpha’s matting refinement. employed to balance performance and efficiency.

\subsection{Mask-Guided Image Encoder}
\label{sec:encoder}
We use a self-supervised pretrained ViT as our image feature extractor, leveraging its strong capacity for feature extraction. To address the inherent limitations of vanilla ViTs in capturing multi-scale and spatial-semantic information, we incorporate a ViT-adapter~\cite{chen2022vitadapter} framework, enabling more comprehensive feature representation. The superior capacity of self-supervised pretrained ViT in encoding image features significantly contributes to the improved performance of our matting approach.

To enable the network to be effectively guided by masks for the purpose of extracting specific content, common approaches, such as the mask attention mechanism in Mask2Former~\cite{cheng2022mask2former} and the Soft-Masked Attention technique in HODOR~\cite{athar2022hodor}, tend to excessively focus on foreground regions. While this focus can enhance foreground extraction in segmentation tasks, applying these methods to image matting may lead to the loss of essential edge details, impacting the quality of fine boundary preservation.

To address this limitation, we propose a Mask-Guided Feature Selection Module(MGFSM) that maintains the network's ability to emphasize foregrounds while preserving edge sensitivity. We construct the mask in a trimap-like manner, assigning distinct values to different regions to provide semantic guidance. 

In the final block of the self-supervised ViT, we apply operations to the self-supervised multi-head attention mechanism, guiding the attention toward region-specific features using a region-specific attention matrix \( \mathbf{S} \). This matrix is defined as:

\[
\mathbf{S} = (\beta \cdot \mathbf{M} + r) \odot (\mathbf{Q} \mathbf{K}^{T}_{\neg \text{[CLS]}}),
\]

where \( \beta \) and \( r \) are scalar parameters, and \( \mathbf{Q} \mathbf{K}^{T}_{\neg \text{[CLS]}} \) represents the attention weights, computed by excluding the [CLS] token. The matrix \( \mathbf{M} \) is a region relevance matrix, which combines with \( \beta \) and \( r \) to apply weighted alignment, ensuring that the attention mechanism focuses more effectively on relevant features in different regions. The matrix \( \mathbf{M} \) is constructed based on region relevance:
\[
\mathbf{M} =
\begin{cases}
    1, & \text{if edge region,} \\
    2, & \text{if foreground,} \\
    0.5, & \text{if background,}
\end{cases}
\]
The adjusted attention output at the last layer is given by:
\[
\text{Attention}(\mathbf{Q}, \mathbf{K}, \mathbf{V}) = \text{softmax}\left(\frac{S +\mathbf{Q}\mathbf{K}^\top}{\sqrt{C}}\right)\mathbf{V}.
\]
 This layer selectively amplifies relevant features, enhancing region-specific detail while preserving the [CLS] token’s integrity.

\subsection{Iterate Decoding}
\label{sec:iter}
Directly predicting alpha values in a single step increases the complexity, as the model must handle both coarse structures and intricate details simultaneously, leading to suboptimal performance in challenging regions.

Inspired by the non-autoregressive generative image transformer used in models like MaskGIT~\cite{maskgit}, we introduce a multi-stage alpha prediction framework that progressively refines the alpha matte over several iterations. This iterative approach simplifies the generation process by breaking it down into manageable steps, allowing the model to incrementally improve the quality of the alpha matte. Each stage fine-tunes the output of the previous iteration, enabling the model to gradually enhance both global context and local details, ultimately producing more accurate alpha predictions.

In our proposed method, we initialize the input with image features and an initial mask, aiming to iteratively refine the alpha predictions by guiding the coarse mask \(\mathbf{M}_{coarse}\) toward a fine-grained alpha matte \(\mathbf{M}_{fine}\). To control the accuracy of each refinement step, we define a confidence score map \( p_\theta(m_{t}^{i,j}|m_{t-1}^{i,j}) \) that predicts the confidence score of the current results. At each step, we selectively sample high-confidence points from this map, which correspond to elements to be transferred. These sampled high-confidence points serve as the input for the next iteration, driving the transition from coarse to refined alpha prediction.

 In our framework, we implement a state transition approach and represent the mask and alpha as two discrete states, initially labeled as \(x_{0}\) and \(x_{T}\), respectively. This allows us to record and control each state transition step, ensuring that the refinement process is stable and progresses unidirectionally toward a fixed outcome.

Formally, let each pixel’s state at timestep \( t \) be denoted by a binary variable \( x_t^{i,j} \), where \( x^{i,j}_0 = [1, 0] \) (representing the coarse state) and \( x^{i,j}_T = [0, 1] \) (representing the fine state) for pixel \((i, j)\). Each refinement step is described by a transition matrix \( \mathbf{Q}_t \), defined as:
\[
\mathbf{Q}_t = 
\begin{bmatrix}
\beta_t & 1 - \beta_t \\
0 & 1 \\
\end{bmatrix},
\]
This matrix ensures that each pixel has a probability \( \beta_t \) of transitioning from the coarse to the fine state, while those already in the coarse state remain unchanged. The forward process is therefore formulated as:
\[
q(x_{t}^{i,j}|x_{t-1}^{i,j}) = x_{t-1}^{i,j} \mathbf{Q}_{t},
\]
By utilizing this unidirectional transition process, we ensure that the refinement of each pixel ultimately converges to the fine state, resulting in a stable alpha matte output aligned with the original mask guidance. This design also mitigates randomness, guaranteeing consistent and precise refinement across iterations.

To enable the network to effectively quantify its prediction confidence, we introduce a confidence loss term, designed to supervise the network in producing confidence scores for the predicted alpha values. The loss function measures the discrepancy between the true and predicted values, guiding the network to reflect this difference in its confidence estimation. Specifically, it is defined as:
\[
\mathcal{L}_C = \left| c_{\alpha} - \left| \alpha_i^p - \alpha_i^g \right| \right|_1,
\]
where \( c_{\alpha} \) represents the confidence score output for each pixel \( i \), and \( \alpha_i^p \) and \( \alpha_i^g \) denote the predicted and ground truth alpha values at pixel \( i \), respectively. This formulation leverages an \( L_1 \) penalty, providing a direct measure of alignment between confidence and prediction accuracy.
\subsection{Self-Guided Sparse Detail Recovery}
\label{sec:decoder}
While the iterate decoding method generates more finegrained alpha matte, it demands substantial memory and computation. To deal with this issue, we introduce the Self-Guided Sparse Detail Recovery (SGSDR) module, building on the foundation of the Sparse High-resolution Module\cite{sun2023sparsemat} (SHM). The SHM selectively enhances details by activating pixels in sparse convolutions, guided by a sparsity map derived from the low-resolution alpha matte \( \mathbf{A}_{\text{l}} \). However, SHM faces the challenge of determining the optimal active regions: the threshold is sensitive, as a value that is too large increases computational costs, while a value that is too small risks losing critical details.

To overcome this limitation, SGSDR introduces a confidence map \( \mathbf{C} \), which adaptively identifies regions requiring refinement. This confidence map is derived from the iterative decoding process and provides a confidence score \( \mathbf{C}(i, j) \) for each pixel. The confidence score is low in areas where the low-resolution output is uncertain, guiding SGSDR to prioritize these regions for further refinement. By leveraging this confidence map, SGSDR can focus on the most challenging areas, improving the overall accuracy and quality of the matting results.

Using this confidence map, SGSDR creates an adaptive sparsity map \( \mathbf{M}_{\text{SG}} \):
\[
\mathbf{M}_{\text{SG}}(i, j) = 
\begin{cases} 
      1, & \text{if } \mathbf{C}(i, j) > \tau,\\
      0, & \text{otherwise},
   \end{cases}
\]
where \( \tau \) is a threshold that controls the activation region, balancing detail recovery with computational efficiency.

With \( \mathbf{M}_{\text{SG}} \) as input, SGSDR performs sparse convolution operations on selected pixels to recover details in the high-resolution alpha matte \( \mathbf{A}_{\text{h}} \). This process can be represented as:
\[
\mathbf{A}_{\text{h}} = \text{SparseResNet}(\mathbf{A}_{\text{l}}, \mathbf{M}_{\text{SG}}),
\]
By incorporating the confidence-guided \( \mathbf{M}_{\text{SG}} \), SGSDR achieves effective detail recovery with reduced computational load, focusing refinement on areas that most benefit from high-resolution enhancement. This approach yields an optimized high-resolution alpha matte while maintaining efficiency.

\begin{algorithm}[t]
\small
\caption{\small Mask2Alpha Training}
\label{algo:train}
\definecolor{codeblue}{rgb}{0.25,0.5,0.5}
\definecolor{codegreen}{rgb}{0,0.6,0}
\definecolor{codekw}{RGB}{207,33,46}
\lstset{
  backgroundcolor=\color{white},
  basicstyle=\fontsize{7.5pt}{7.5pt}\ttfamily\selectfont,
  columns=fullflexible,
  breaklines=true,
  captionpos=b,
  commentstyle=\fontsize{7.5pt}{7.5pt}\color{codegreen},
  keywordstyle=\fontsize{7.5pt}{7.5pt}\color{codekw},
  escapechar={|}, 
}
\begin{lstlisting}[language=python]
def train(F, M_fine, M_coarse, T):
  """
  F: image features, M_fine: ground truth alpha
  M_coarse: coarse mask, T: total iteration steps
  """

  # Initialize mask and binary state
  m_0 = M_fine
  x_0 = [1] # Binary mask: 1 represents the fine state

  # Sample time step and transition matrix
  t = uniform(0, 1)     # time step
  q = schedule_q(t)      # generate transition matrix

  # Sample and apply pixel transition
  x_t = sample(q)        # sample from transition matrix
  m_t = x_t * M_coarse + (1 - x_t) * M_fine

  # Decode and compute loss
  alpha_predict, transition_probability = iterate_decoder(F, m_t, x_t)
  loss = compute_loss(alpha_predict, M_fine)

  # Update parameters
  update_parameters(loss)
\end{lstlisting}

\end{algorithm}

\subsection{Training and Inference for Iterate Decoding}
\label{sec:procedure}

\paragraph{Training}
The training process is structured as follows (see Algorithm \ref{algo:train}). We start with the total number of iteration steps \( T \) and a dataset \( \mathcal{D} = \{(\mathcal{F}, \mathcal{M}_{fine}, \mathcal{M}_{coarse})^K\} \). During each iteration, we sample a training example \( (\mathcal{F}, \mathcal{M}_{fine}, \mathcal{M}_{coarse}) \) from the dataset and randomly select an iteration step \( t \) uniformly from the range \( 1 \) to \( T \). We initialize the mask \( m_0 \) with \( M_{fine} \) and set the transition variable \( x_0^{i,j} \) to \([1]\).

Next, we compute the transition probability \( q(x_{t}^{i,j}|x_{0}^{i,j}) \) and sample \( x_{t}^{i,j} \) from this distribution, yielding a binary representation \( x_t \in \{0, 1\}^{2 \times H \times W} \). The transition of pixels is then determined using the formula:
\[
m_t = x_t[0] \odot \mathbf{M}_{fine} + x_t[1] \odot \mathbf{M}_{coarse},
\]

Finally, we perform a gradient descent step to optimize the loss function:
\[
\nabla_\theta \mathcal{L}(f_\theta(I, m_t, t), \mathbf{M}_{fine}),
\]

This process continues until convergence is achieved.

\begin{algorithm}[t]
\small
\caption{\small Mask2Alpha Inference}
\label{algo:inference}
\definecolor{codeblue}{rgb}{0.25,0.5,0.5}
\definecolor{codegreen}{rgb}{0,0.6,0}
\definecolor{codekw}{rgb}{0.85, 0.18, 0.50}
\lstset{
  backgroundcolor=\color{white},
  basicstyle=\fontsize{7.5pt}{7.5pt}\ttfamily\selectfont,
  columns=fullflexible,
  breaklines=true,
  captionpos=b,
  commentstyle=\fontsize{7.5pt}{7.5pt}\color{codegreen},
  keywordstyle=\fontsize{7.5pt}{7.5pt}\color{codekw},
  escapechar={|}, 
}
\begin{lstlisting}[language=python]
def infer(F, M_coarse, T):
  """
  F: image features from image encoder
  M_coarse: coarse mask, T: total iteration steps
  """
  
  # Initialize binary mask, 0 represents the coarse state
  x_t = [0]
  m_t = M_coarse
  
  for t in range(T, 0, -1):
    # Predict alpha and confidence map for current transition state
    confid_alpha, pred_alpha = iterate_decoder(F, m_t, x_t)
    
    # Sample transition state
    sample_xt = sample(confid_alpha)
    
    # Update current state and transition
    current_xt = update(sample_xt, x_t)
    m_t = current_xt * pred_alpha + (1 - current_xt) * m_t

  return pred_alpha  # Return predict alpha
\end{lstlisting}

\end{algorithm}

\paragraph{Inference.}
The inference procedure, detailed in Algorithm \ref{algo:inference}, begins by initializing the transition variable \( \mathbf{x}_T = [0] \) and setting the mask \( \mathbf{m}_T = \mathbf{M}_{coarse} \). For each iteration \( t \) from \( T \) down to \( 1 \), the process involves predicting and refining the mask based on the following steps.

First, the predicted fine mask \( \tilde m_{0|t} \) and its probability \( p_\theta(\tilde m_{0|t}) \) are computed using the function \( f_\theta(I, m_t, t) \), which incorporates image features, the current mask \( m_t \), and time step \( t \). 

Next, the transition probability \( p_\theta(x_{t-1}^{i,j}|x_{t}^{i,j}) \) is defined based on the current state, guiding the sampling of \( x_{t-1}^{i,j} \) from this transition distribution, resulting in the binary variable \( x_t \in \{0, 1\}^{2 \times H \times W} \).

The mask \( m_{t-1} \) is then updated by combining \( \tilde m_{0|t} \) and \( \mathbf{M}_{coarse} \) through:
\[
m_{t-1} = x_{t-1}[0] \odot \tilde{\mathbf{m}}_{0|t} + x_{t-1}[1] \odot \mathbf{M}_{\text{coarse}},
\]

After completing all iterations, the refined mask \( m_0 \) is returned. This procedure gradually transitions the mask to higher levels of detail while leveraging both the coarse mask and fine-grained predictions for optimal refinement.

 % Input latest version
    \else
        \input{sec/3_method_v4} % Input old version with suffix
    \fi

\section{Experiments}\label{sec:exp}

\begin{table*}[t]
\footnotesize
    \centering
    \setlength{\tabcolsep}{1.8mm}{
    \begin{tabular}{l|c|c|c|c|c|c|c|c|c|c|c|c|c}
        \hline\hline
        \multicolumn{1}{c|}{\multirow{2}{*}{Method}} &
        \multicolumn{1}{c|}{\multirow{2}{*}{Category}} &
        \multicolumn{4}{c|}{\multirow{1}{*}{AIM-500}} &
        \multicolumn{4}{c|}{\multirow{1}{*}{AM2K}} &
        \multicolumn{4}{c}{\multirow{1}{*}{P3M-500-NP}} 
        \\
        \cline{3-14}
        \multicolumn{1}{c|}{} & \multicolumn{1}{c|}{} &
        \multicolumn{1}{c}{SAD$\downarrow$} &
        \multicolumn{1}{c}{MSE$\downarrow$} & 
        \multicolumn{1}{c}{Grad$\downarrow$} & 
        \multicolumn{1}{c|}{Conn$\downarrow$} & 
        \multicolumn{1}{c}{SAD$\downarrow$} &
        \multicolumn{1}{c}{MSE$\downarrow$} & 
        \multicolumn{1}{c}{Grad$\downarrow$} & 
        \multicolumn{1}{c|}{Conn$\downarrow$} &
        \multicolumn{1}{c}{SAD$\downarrow$} &
        \multicolumn{1}{c}{MSE$\downarrow$} & 
        \multicolumn{1}{c}{Grad$\downarrow$} & 
        \multicolumn{1}{c}{Conn$\downarrow$} 
        \\
        \hline
        P3M-ViTAE~\cite{rethink_p3m} & Human & 111.22 & 0.0595 & 44.16 & 54.02 & 40.34 & 0.0205 & 38.71 & 20.55 & \textbf{8.88} & 0.0023 & \underline{8.33} & \underline{11.22} \\
GFM~\cite{liu2024aematter} & Animal & 95.50 & 0.0503 & 74.38 & 46.97 & \textbf{11.18} & 0.0031 & \textbf{10.27} & \textbf{9.77} & 110.80 & 0.0606 & 106.28 & 33.97 \\
AIM~\cite{sun2021sim} & Natural & 48.73 & 0.0187 & 47.96 & \underline{34.75} & 28.13 & 0.0102 & 26.89 & 19.25 & 29.46 & 0.0114 & 28.51 & 25.85  \\
\hline

        % ViTMatte-B~\cite{yao2024vitmatte} & Mixed & & & & & & & & & & & & 
        % \\
    MGM~\cite{yu2021mgm} & Natural            & 51.82 & 0.0126 & 33.18 & 51.78 & 22.69 & 0.0039 & 13.57 & 21.37 & 15.35 & 0.0025 & 14.67 & 14.53 \\
    % SparseMat~\cite{sun2023sparsemat} & Human & 45.54&0.0187&45.33&43.29 & 25.03 & 0.0055 & 22.16 & 24.94 & 20.05&0.0075&0.0019&23.9 \\
    MaGGIe~\cite{huynh2024maggie} & Human     & 47.65 & 0.0121 & 37.31 & 45.97 & 16.59 & 0.0026 & 12.49 & 15.82 & 11.39 & \underline{0.0017} & 13.52 & \textbf{10.86} \\
    MGM$^\dagger$~\cite{park2023mgmwild}& Natural & \underline{43.05} & \underline{0.0102} & \underline{32.13} & 42.71 & 17.23 & \underline{0.0024} & 12.71 & 16.08 & 13.77 & 0.0021 & 15.27 & 13.08 \\
    \textbf{Mask2Alpha}(ours) & Natural & \textbf{35.61} & \textbf{0.0091} & \textbf{29.74} & \textbf{31.07} & \underline{13.22} & \textbf{0.0021} & \underline{10.55} & 
    \underline{10.37} & \underline{9.84} & \textbf{0.0015} & \textbf{8.03} & 12.07 \\

        \hline\hline       
    \end{tabular}}
    \vspace{-0.1in}
    \caption{{\bf Quantitative Comparisons Across Diverse Real-World Datasets.} Metrics include SAD, MSE, Grad, and Conn. lower values indicate better performance. Bold numbers indicate the best performance.}
    \label{tab:main}
    \vspace{-0.1in}
\end{table*}

\begin{figure*}
    \centering
    \includegraphics[width=1.0\linewidth]{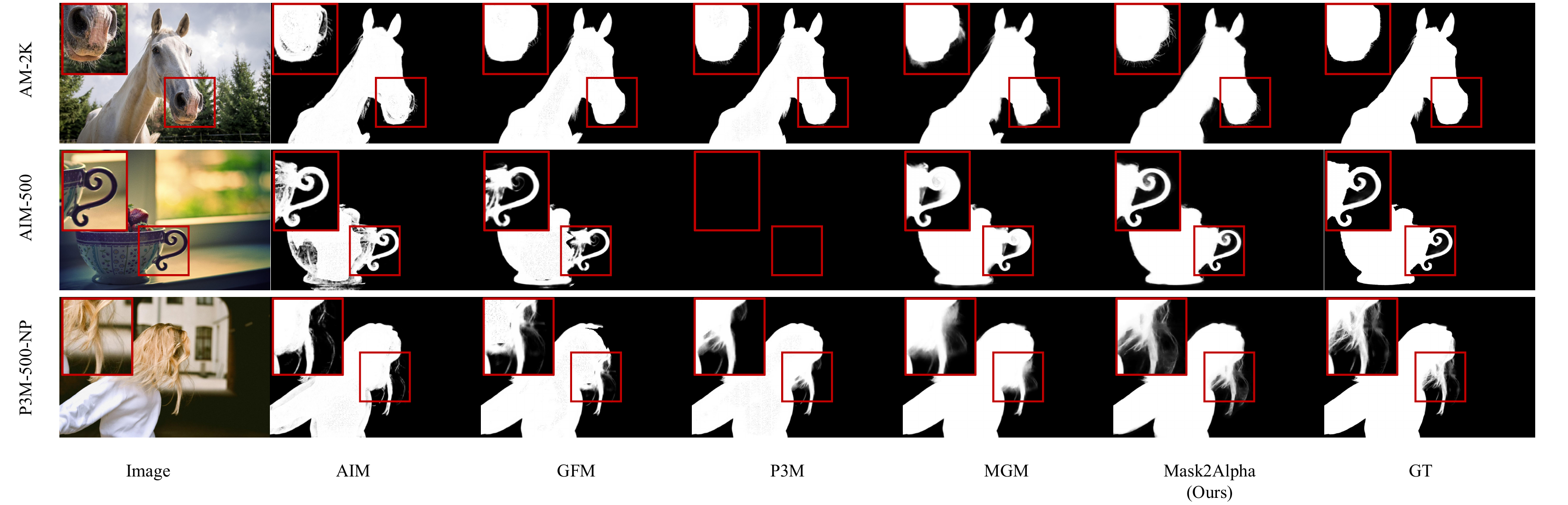}
    \vspace{-0.25in}
    \caption{{\bf Qualitative Comparisons Across Diverse Real-World Datasets.} Our method demonstrates superior generalization ability across various category-diverse real-world datasets, surpassing category-specific models. It shows enhanced semantic understanding, and improved detail-handling capability in complex scenes compared to mask-guided methods.}
    \label{fig:mainres}
    \vspace{-0.2in}
\end{figure*}

\begin{table}[t]
    \centering
    \setlength{\tabcolsep}{2.0mm}{
    \begin{tabular}{l|c|c|c|c}
    \hline\hline
         Method & SAD & MSE & Grad & Conn \\
         \hline
     MGM$^\dagger$~\cite{park2023mgmwild} &28.68 &0.0034 &15.07 &26.19\\
     MaGGIe~\cite{huynh2024maggie} &\underline{20.01} &\underline{0.0021}
     &12.53 &17.49\\   InstMatt~\cite{sun2023sparsemat} &\textbf{19.48} &\textbf{0.0017} &\underline{11.61}
     &\textbf{17.02}\\ 
         \hline
     Mask2Alpha(ours) &20.35 &\underline{0.0021} &\textbf{11.28} &\underline{17.08}\\
         \hline\hline
    \end{tabular}}
    \vspace{-0.1in}
    \caption{{\bf Quantitative Comparison with Instance-Aware Methods.} Unlike instance-aware methods that are trained on multi-instance human-specific datasets, our approach has not been trained on such datasets yet still demonstrates strong competitiveness in instance awareness across diverse scenarios.}
    \label{tab:instance_aware}
    \vspace{-0.1in}
\end{table}

In this chapter, we provide a comprehensive evaluation of our proposed Mask2Alpha framework, detailing the experimental setup in Section~\ref{sec:detail}, we describe the datasets, evaluation metrics, and training details used in our experiments.
In Section~\ref{sec:main}, we compare Mask2Alpha's performance against state-of-the-art methods and evaluate its generalization across various datasets. In Section~\ref{sec:ablat}, we conduct ablation studies to demonstrate the contributions of key components to overall performance, with detailed results provided in the appendix.

\subsection{Implementation Details}
\label{sec:detail}
{\bf Datasets.}
For training, we only use the DIM\cite{xu2017dim} and COCO\cite{lin2014coco} datasets. To rigorously assess our model's generalization across different domains, we evaluate it on several distinct natural datasets, including AIM-500\cite{li2021aim500} for natural images, P3M-500-NP\cite{rethink_p3m} for human segmentation, and AM-2K\cite{li2022am2k} for animal segmentation. These datasets allow us to examine our model’s accuracy and versatility across various foreground types. To assess instance recognition and robustness with different mask types, we include evaluations on the M-HIM2K~\cite{huynh2024maggie} dataset, which provides high-quality, instance-specific human masks.

\noindent
{\bf Evaluation Metrics.}
We employ four widely recognized metrics for evaluating image matting performance: Sum of Absolute Differences (SAD), Mean Squared Error (MSE), Gradient (Grad), and Connectivity (Conn). For each metric, lower values indicate better performance. 

\noindent
{\bf Training Details.}
Following previous works\cite{park2023mgmwild,yu2021mgm,xu2017dim,Lu2019indices}, we utilize DIM and COCO as the matting and background datasets for our experiments. Specifically, we leverage pre-trained weights from ViT-Adapter\cite{chen2022vitadapter} and BEiTv2\cite{peng2022beitv2} as the image encoder, which provide strong feature extraction capabilities. To accommodate high-resolution image processing, we set the cropping size to 1024 pixels.

\subsection{Main Results}
\label{sec:main}

We evaluate our model across three key dimensions: generalization ability across object types in natural matting, instance-awareness for distinguishing multiple objects, and matting capability in complex real-world scenarios.

\noindent
{\bf Generalization in Natural Matting.} To evaluate the generalization capability of our model across various object categories, we benchmark it against three representative category-specific matting methods: P3M-ViTAE\cite{rethink_p3m} for human matting, GFM\cite{li2022am2k} for animals, and AIM\cite{li2021aim500} for natural scenes. The evaluation spans three datasets aligned with these categories: P3M-500-NP for human portraits, AM2K for animals, and AIM-500 for natural images. We further compare our model with several state-of-the-art mask-guided approaches, including MaGGIe\cite{huynh2024maggie}, MGM\cite{yu2021mgm}, and MGM-in-the-Wild\cite{park2023mgmwild}. For clarity, we denote MGM-in-the-Wild as MGM$^\dagger$ hereafter, as we re-implemented this model following its original training setup.

Our model, trained solely on the DIM dataset, demonstrates strong generalization across domains without any fine-tuning for specific categories. As shown in Table \ref{tab:main}, it achieves high-quality results across all object categories. In contrast, category-specific methods perform best within their respective domains but show limitations in others, as evidenced in the qualitative results shown in Figure \ref{fig:mainres}. Additionally, compared to other mask-guided methods, our approach achieves state-of-the-art performance and performs significantly better than MGM$^\dagger$, which is tailored specifically for in-the-wild scenarios.

\noindent
{\bf Instance Awareness and Differentiation.} To evaluate our model's capability in distinguishing multiple objects within a scene, we conduct experiments on the natural subset of the M-HIM2K\cite{huynh2024maggie} dataset, using segmentation results from the R50-C4-3x\cite{cheng2022mask2former} model as guidance. Unlike instance-aware methods trained specifically on multi-instance human datasets, our approach has not been trained on such specialized datasets. Nonetheless, as shown in Table \ref{tab:instance_aware}, 
our model demonstrates strong instance-awareness, effectively identifying and separating overlapping objects, and maintains competitive performance across diverse scenarios.

\noindent
% {\bf Matting in Complex Real-World Scenarios.} Lastly, we evaluate our model’s matting capability in complex real-world environments using the COCO dataset. The COCO dataset introduces diverse and intricate backgrounds, testing our model’s resilience to complex occlusions and background noise. As illustrated in Figure , our model performs consistently well, generating accurate and continuous alpha mattes. These results underscore our model’s adaptability to varied real-world conditions, enabling reliable matting performance in unpredictable scenarios.

\subsection{Ablation Study}
\label{sec:ablat}
We conduct ablation studies on the three main modules in our Mask2Alpha framework, using AIM-500 as the default dataset unless otherwise specified.Due to space limitations, we have placed some qualitative experiments in the supplementary material.

\noindent
{\bf Analysis of Mask-Guided Feature Selection Module.} This module enhances the model’s capability to differentiate instances effectively. To evaluate its impact, we visualize instance selection results on the M-HIM2K dataset and compare performance with and without the module. The results demonstrate significant improvements in instance differentiation, as shown in Table \ref{tab:MGFSM_ablation}.

\begin{table}[t]
    \centering
    \setlength{\tabcolsep}{2.5mm}{
    \begin{tabular}{l|c|c|c|c}
    \hline\hline
         Method & SAD & MSE & Grad & Conn \\
         \hline
         w/o MGFSM & 45.90 & 0.0178 & 42.94 & 47.82 \\
         w. MGFSM & 36.05 & 0.0121 & 29.80 & 34.12 \\
         \hline\hline
    \end{tabular}}
    \vspace{-0.1in}
    \caption{Ablation results of Mask-Guided Feature Selection Module.}
    \label{tab:MGFSM_ablation}
    \vspace{-0.1in}
\end{table}

\noindent
{\bf Analysis of Iterative Decoding.} We investigate the effect of varying iteration counts within the iterative decoding process in our Mask2Alpha framework. Specifically, we conduct ablation studies with different decoding iterations to assess how iteration count impacts output quality. As shown in Table \ref{tab:iter_steps}, increasing the number of iterations consistently refines the details of the final output, with optimal performance achieved at N iterations.

% Additionally, we evaluate the impact of the state transition matrix on the stability of the decoding process. As shown in Table \ref{tab:state_trans}, by introducing structured transition states, the matrix reduces unpredictable variations across iterations, resulting in more stable and reliable outputs. This stability contributes to improved overall results, allowing the model to progressively enhance detail while minimizing erratic changes. These findings underscore the combined importance of iterative refinement and the state transition matrix in achieving high-quality and dependable predictions.

\begin{table}[t]
    \centering
    \setlength{\tabcolsep}{2.5mm}{
    \begin{tabular}{l|c|c|c|c|c}
    \hline\hline
         Steps \( N \) & None & 3 & 6 & 10 & 20 \\
         \hline
         SAD & 37.98 & 37.80 & 36.83 & 36.54 & 36.53 \\
         \hline
         Time (s)& n/a & 0.030 & 0.054 & 0.087& 0.167 \\
         \hline\hline
    \end{tabular}}
    \vspace{-0.1in}
    \caption{Impact of Iteration Steps \( t \) on the Accuracy and Efficiency of Mask2Alpha.}
    \label{tab:iter_steps}
    \vspace{-0.1in}
\end{table}

% \begin{table}[t]
%     \centering
%     \setlength{\tabcolsep}{1mm}{
%     \begin{tabular}{l|c|c|c|c}
%     \hline\hline
%          Method & SAD & MSE & Grad & Conn \\
%          \hline
%          w/o state translation matrix  \\
%          w. state translation matrix  \\
%          \hline\hline
%     \end{tabular}}
%     \vspace{-0.1in}
%     \caption{Ablation results of state translation matrix.}
%     \label{tab:state_trans}
%     \vspace{-0.1in}
% \end{table}

\begin{figure}[t]
    \centering
    \includegraphics[width=1\linewidth]{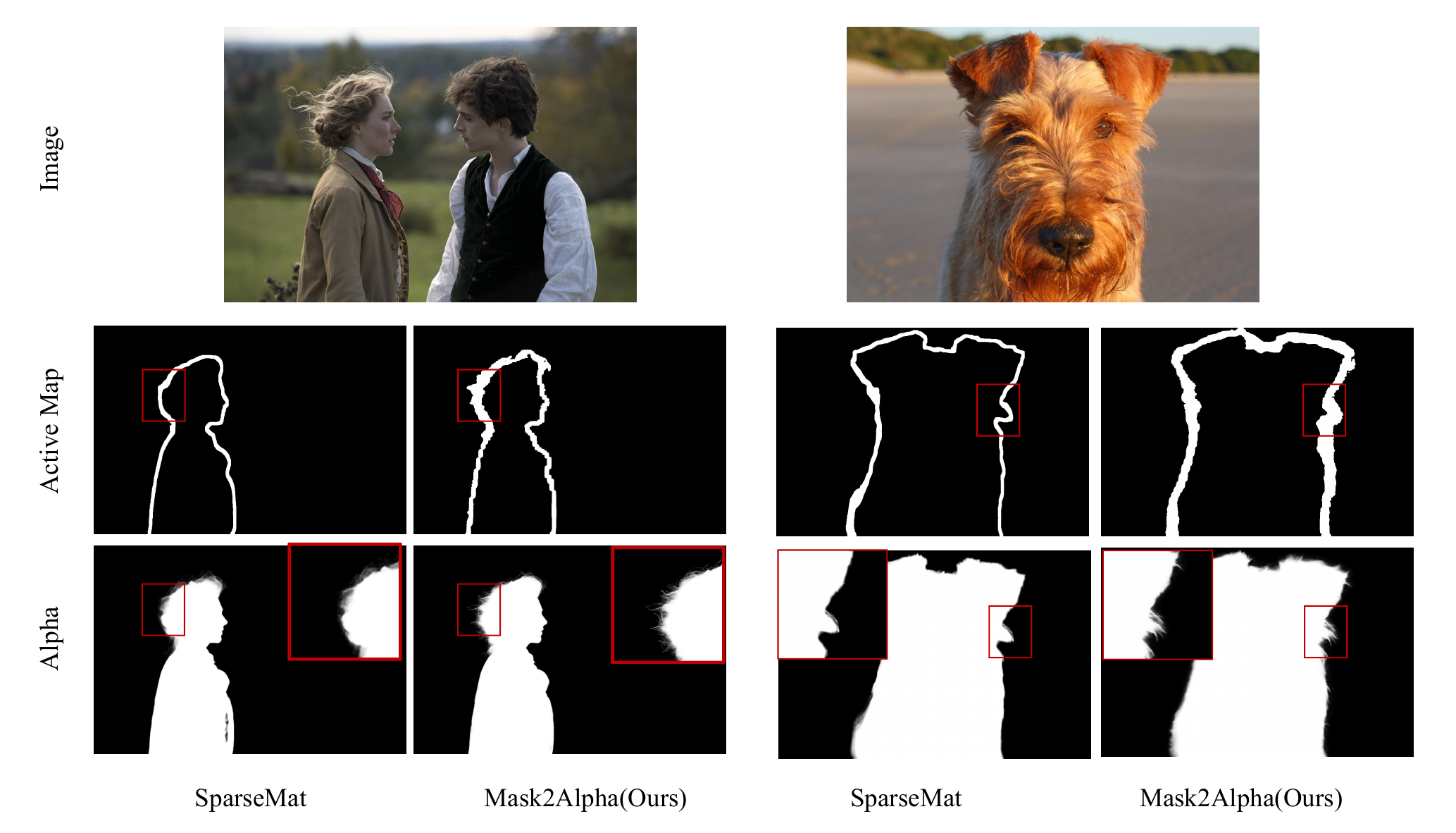}
    \caption {Qualitaive results of sparse activation maps. The second row presents sparse activation maps, comparing our method and Sparsemat\cite{sun2023sparsemat}. Our self-guided approach automatically activates more regions based on fine-grained details.}
    \label{fig:activate}
\end{figure}

\noindent
{\bf Analysis of Self-Guided Sparse Detail Recovery.} This module is compared with the SparseMat method by visualizing activation indices and assessing the ability to selectively recover high-resolution details. As illustrated in Figure~\ref{fig:activate}, our self-guided approach automatically activates more regions based on fine details, especially around boundary areas in synthetic and real scenes. Visual results show that our method captures finer details more accurately. Additionally, Table~\ref{tab:sgsdr_ablation} compares our module with an alternative approach that incorporates the SparseMat module into our method, demonstrating that our approach achieves better performance while preserving computational efficiency.

\begin{table}[t]
    \centering
    \setlength{\tabcolsep}{1.5mm}{
    \begin{tabular}{l|c|c|c|c}
    \hline\hline
         Method & SAD & MSE & Grad & Conn \\
         \hline
         Baseline & 45.87 & 26.68 & 64.70 & 76.02 \\
        Baseline + SHM & 40.40 & 24.38 & 42.94 & 77.82 \\
         Baseline + SGSDR(Ours)& 37.05 & 19.98 & 28.80 & 53.47 \\
         \hline\hline
    \end{tabular}}
    \vspace{-0.1in}
    \caption{Comparison of Mask2Alpha with Self-Guided Sparse Detail Recovery (SGSDR) or SHM, where the baseline represents the result without detail recovery at low resolution.}
    \label{tab:sgsdr_ablation}
    \vspace{-0.1in}
\end{table}

\section{Conclusion}
In this work, we introduced Mask2Alpha, a novel framework designed to overcome the challenges of image matting. Our extensive experiments demonstrate that Mask2Alpha significantly outperforms existing methods, particularly in complex and cluttered environments. By integrating iterative refinement and instance-aware processing, our approach effectively addresses the challenges of matting in complex scenes, particularly in handling fine details and object boundaries. The integration of a mask-guided feature selection module enhances the model's ability to distinguish between multiple instances, which is crucial for matting in complex scenes. Additionally, the use of self-supervised ViT features allows the model to capture high-level contextual information, further improving its performance in diverse and semantically rich scenarios. The lightweight decoder effectively fuses low-level visual details with semantic understanding, ensuring computational efficiency without compromising performance.
{
    \small
    \bibliographystyle{ieeenat_fullname}
    \bibliography{cite}
}

% \input{supplementary}

\end{document}